\documentclass[letterpaper, 10 pt, conference]{ieeeconf}  

\IEEEoverridecommandlockouts                              

\usepackage{myStyle}

\title{\LARGE \bf
Active Disturbance Rejection Control for Trajectory Tracking of a Seagoing USV: Design, Simulation, and Field Experiments
}

\author{Jelmer van der Saag$^{1,2}$, Elia Trevisan$^1$, Wouter Falkena$^2$ and Javier Alonso-Mora$^1$
\thanks{This research is supported by the project ``Sustainable Transportation and Logistics over Water: Electrification, Automation and Optimization (TRiLOGy)'' of the Netherlands Organization for Scientific Research (NWO), domain Science (ENW), and the Amsterdam Institute for Advanced Metropolitan Solutions (AMS) in the Netherlands.}
\thanks{$^{1}$Cognitive Robotics Department,
        TU Delft, 2628 CD Delft, The Netherlands
        {\tt\small, j.alonsomora\}@tudelft.nl}}
\thanks{$^{2}$Demcon Unmanned Systems, 
        Den Haag, 2583 WC, the Netherlands.}%
}
\DeclareAcronym{ADRC}{
    short = ADRC,
    long = Active Disturbance Rejection Control
}

\DeclareAcronym{TD}{
    short = TD,
    long = Tracking Differentiator
}

\DeclareAcronym{NLSEF}{
    short = NLSEF,
    long = Nonlinear Error State Feedback
}

\DeclareAcronym{USV}{
    short = USV,
    long = Unmanned Surface Vessel
}

\DeclareAcronym{PID}{
    short = PID,
    long = Proportional-Integral-Derivative
}

\DeclareAcronym{ESO}{
    short = ESO,
    long = Extended State Observer
}

\DeclareAcronym{SISO}{
    short = SISO,
    long = Single-Input Single-Output
}

\DeclareAcronym{MIMO}{
    short = MIMO,
    long = Multi-Input Multi-Output
}

\DeclareAcronym{DUS}{
    short = DUS,
    long = Demcon Unmanned Systems
}

\DeclareAcronym{SMC}{
    short = SMC,
    long = Sliding Mode Control
}

\DeclareAcronym{GNSS}{
    short = GNSS,
    long = Global Navigation Satelite System
}

\DeclareAcronym{IMU}{
    short = IMU,
    long = Inertial Measurement Unit
}

\DeclareAcronym{DOF}{
    short = DOF,
    long = Degrees-of-Freedom
}

\DeclareAcronym{NED}{
    short = NED,
    long = North-East-Down
}

\DeclareAcronym{SITL}{
    short = SITL,
    long = Software-In-The-Loop
}

\DeclareAcronym{DWP2}{
    short = DWP2,
    long = Dynamic Water Physics 2
}

\DeclareAcronym{XTE}{
    short = XTE,
    long = Cross-Track Error
}

\DeclareAcronym{RMS}{
    short = RMS,
    long = Root Mean Square
}

\DeclareAcronym{ILOS}{
    short = ILOS,
    long = Integral Line-of-Sight
}

\DeclareAcronym{RPM}{
    short = RPM,
    long = Rotations Per Minute
}%

\begin{document}

\maketitle
\thispagestyle{empty}
\pagestyle{empty}

\begin{abstract}
Unmanned Surface Vessels (USVs) face significant control challenges due to uncertain environmental disturbances like waves and currents. This paper proposes a trajectory tracking controller based on Active Disturbance Rejection Control (ADRC) implemented on the DUS V2500. A custom simulation incorporating realistic waves and current disturbances is developed to validate the controller's performance, supported by further validation through field tests in the harbour of Scheveningen, the Netherlands, and at sea. Simulation results demonstrate that ADRC significantly reduces cross-track error across all tested conditions compared to a baseline PID controller but increases control effort and energy consumption. Field trials confirm these findings while revealing a further increase in energy consumption during sea trials compared to the baseline. 
Videos can be found at \href{https://autonomousrobots.nl/paper_websites/adrc-demcon}{https://autonomousrobots.nl/paper\_websites/adrc-demcon}.
\end{abstract}


\section{Introduction}
%
%
%
%
Although oceans cover two-thirds of the Earth's surface and approximately 37\% of the global population resides within 100 kilometres of a shoreline \cite{marinerobot_applications}, the ocean floor remains largely unmapped with only 15\% coverage as of July 2019 \cite{oceanfloor_coverage}. Although still largely unexplored, the maritime domain remains critical for economic, scientific, and military advancements \cite{usv_developmentchallengeoverview}, highlighting the need for technological developments in marine exploration and navigation.

One such development has been the introduction of the \ac{USV}. These semi-autonomous vessels typically forgo an on-board crew, relying instead on a remote human operator or no operator at all. This shift significantly reduces crew requirements compared to traditional crewed vessels, resulting in lower operational costs and improved work safety, as an operator can remain in a safe environment \cite{usv_developmentchallengeoverview}. Furthermore, the absence of a crew enables the design of smaller and more energy-efficient vessels, thus reducing their environmental footprint \cite{smallusv_safetyassurance}. These benefits make \acs{USV}s a suitable alternative to conventional vessels for a wide range of applications, including hydrographic surveys, offshore inspections, and maritime exploration \cite{advances_umv_book, usv_offshore_dewinter}.

The maritime environment in which \acs{USV}s operate can be challenging due to environmental factors such as waves and currents. These disturbances are often unpredictable and negatively
impact tasks such as trajectory tracking. Despite these challenges, path-following control has often relied on traditional control methods, and \ac{PID} control remains the most popular approach in recent academic papers on \acs{USV}s for its simplicity and ability to provide satisfactory performance \cite{haiton2023_reviewpathfollowingcontrol}. However, PID control may struggle to maintain performance due to non-linear system dynamics or severe environmental disturbances, loosing performance in challenging environments.

\begin{figure}[t!]
    \centering
    \includegraphics[width=\linewidth]{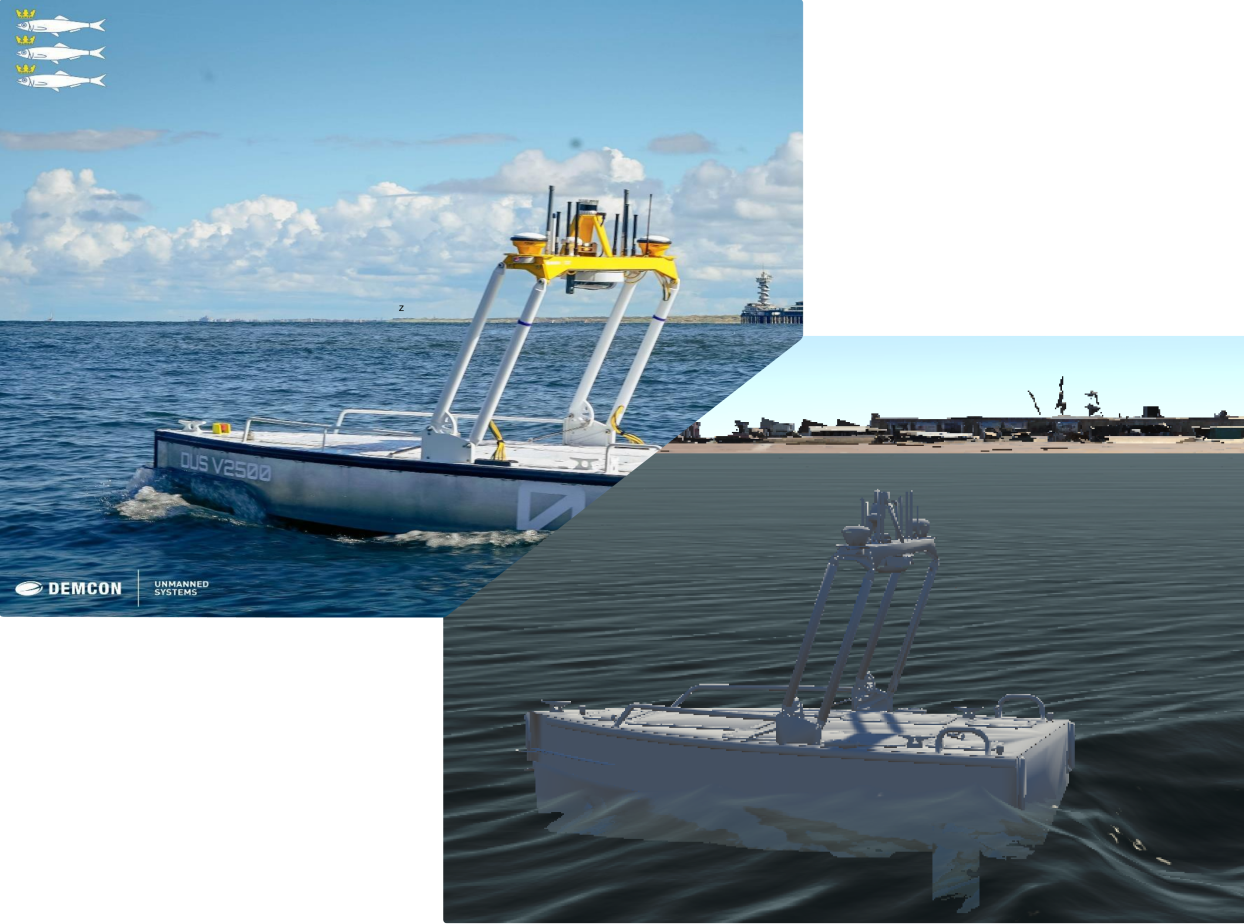}
    \caption{DUS V2500 USV platform, operating near-shore in calm conditions in Scheveningen, the Netherlands (left) \textcopyright \hspace{0.1cm}Demcon Unmanned Systems. and in similar conditions in simulation (right).}
    \label{fig:DUS_V2500}
\end{figure}

Various control strategies have recently been proposed to improve \ac{USV} performance in dynamic environments. Alternatives to \acs{PID}, such as \ac{SMC} \cite{mcninch2009_optimalslidingcontrol}, feedback linearisation \cite{borhaug2011_feedbacklinearisation}, and backstepping control \cite{fossen2003_lineofsightpathfollowing} have been widely applied \cite{haiton2023_reviewpathfollowingcontrol}. While these methods can offer robustness and improved tracking, they require varying degrees of system modelling, and their performance often hinges on the accuracy of such models. In highly uncertain maritime environments, where wave and current disturbances are difficult to reliably model, these approaches may be less practical. Additionally, techniques such as SMC may suffer from chattering effects \cite{zhu2021_modelfreeSMC}, while nonlinear MPC solvers are often too computationally expensive for real-time control \cite{wang2024_policylearningMPC}. Recent advances aimed at reducing this dependency include the use of neural networks to approximate unknown dynamics \cite{qiu2019_USVlinearisationcontrol} and techniques to replace system information with rapid measurements of state and disturbance \cite{wei2023_modelfreesmc_usv}.

One technique aimed at eliminating model dependency, marketed as an extension to \acs{PID}, is \ac{ADRC}. It has seen adoption in USV control due to its ability to estimate uncertain system dynamics and external disturbances without any system information \cite{han2009_frompidtoadrc}. It allows for estimation of the \enquote{total disturbance} acting on the system, including internal uncertainties and external disturbances, which is compensated via a feed-forward component. ADRC was selected in this work because of its ability to maintain performance in the presence of unknown and time-varying disturbances, while not requiring a prior model of the vessel or environment. Compared to SMC or MPC, ADRC is touted as a direct upgrade to PID control \cite{han2009_frompidtoadrc}, serving as a logical upgrade for existing USV platforms that currently use PID controllers, such as the DUS V2500.

Although the effectiveness of \ac{ADRC} in \ac{USV} control has been repeatedly demonstrated in simulation studies \cite{Hu2021_ADRCcoursecontrol, chen2015_controlandsimofADRCofUSV, fan2021_ADRCcoursecontrol_fuzzytuning, Zheng2022_softactorADRC, fu2021_ADRCIPSO, li2022_dpADRC}, real-world validations are limited. Existing experiments have been conducted only in inland waters \cite{Wu2022_watersamplingADRC}, which lack significant disturbances such as stronger waves and currents. To the best of the authors’ knowledge, no prior research has evaluated the performance of \ac{ADRC} implemented on a seagoing \ac{USV} through real-world tests.

We address this gap by designing and validating an \ac{ADRC}-based control strategy on the \acs{DUS} V2500, a fully electric \ac{USV} developed by \acf{DUS} as shown in \autoref{fig:DUS_V2500}. Capable of operating in conditions up to Sea State 4, the \ac{DUS} V2500 is a suitable platform for evaluating \ac{ADRC} performance in near-shore conditions. Performance is evaluated through simulation in a Unity environment featuring realistic waves and currents and field trials in Scheveningen, the Netherlands. Through these tests, we evaluate the feasibility of \ac{ADRC} for seagoing applications. The contributions of this paper are as follows:

\begin{itemize}
    \item The design and implementation of a trajectory-tracking second-order \ac{ADRC} controller tailored for the \ac{DUS} V2500, an underactuated, rudderless \ac{USV}.
    \item The development and verification of a Unity-based simulation environment incorporating realistic wave, current, and wind disturbances to facilitate controller evaluation.
    \item The adaptation of a control strategy from quadrotor systems to a USV, addressing the challenges posed by propeller delay in an underactuated maritime platform.
    \item Real-world experimental validation of an \ac{ADRC} controller for a \ac{USV} at sea.
\end{itemize}

\section{Preliminaries}
\label{sec:preliminaries}
This section provides an overview of the \ac{DUS} V2500 platform, followed by a description of the system dynamics.

\subsection{USV Platform}
The \ac{DUS} V2500, as shown in \autoref{fig:DUS_V2500}, is a fully electric \ac{USV} designed by Demcon Unmanned Systems for applications such as inspection and hydrography. The vessel measures approximately 2.5 metres in length and is primarily intended for inland and near-shore operations. It is rated to operate in conditions up to Douglas Sea State 4, corresponding to wave heights of up to 2.5 metres.

The platform is underactuated, featuring two stern-mounted thrusters and a single bow thruster for manoeuvring. The \ac{DUS} V2500 operates without an onboard crew semi-autonomously, executing missions autonomously based on a predefined mission plan monitored by a remote operator.

The vessel's localisation is achieved through the fusion of \ac{GNSS} and \ac{IMU} data, providing estimates of the pose with an accuracy of $\pm0.01$ metres. A detailed description of the localisation algorithm and system architecture of the \ac{DUS} V2500 remains proprietary to Demcon Unmanned Systems and is beyond the scope of this paper.

\subsection{System Dynamics}
\begin{figure}
    \centering
    \includegraphics[width=0.6\linewidth]{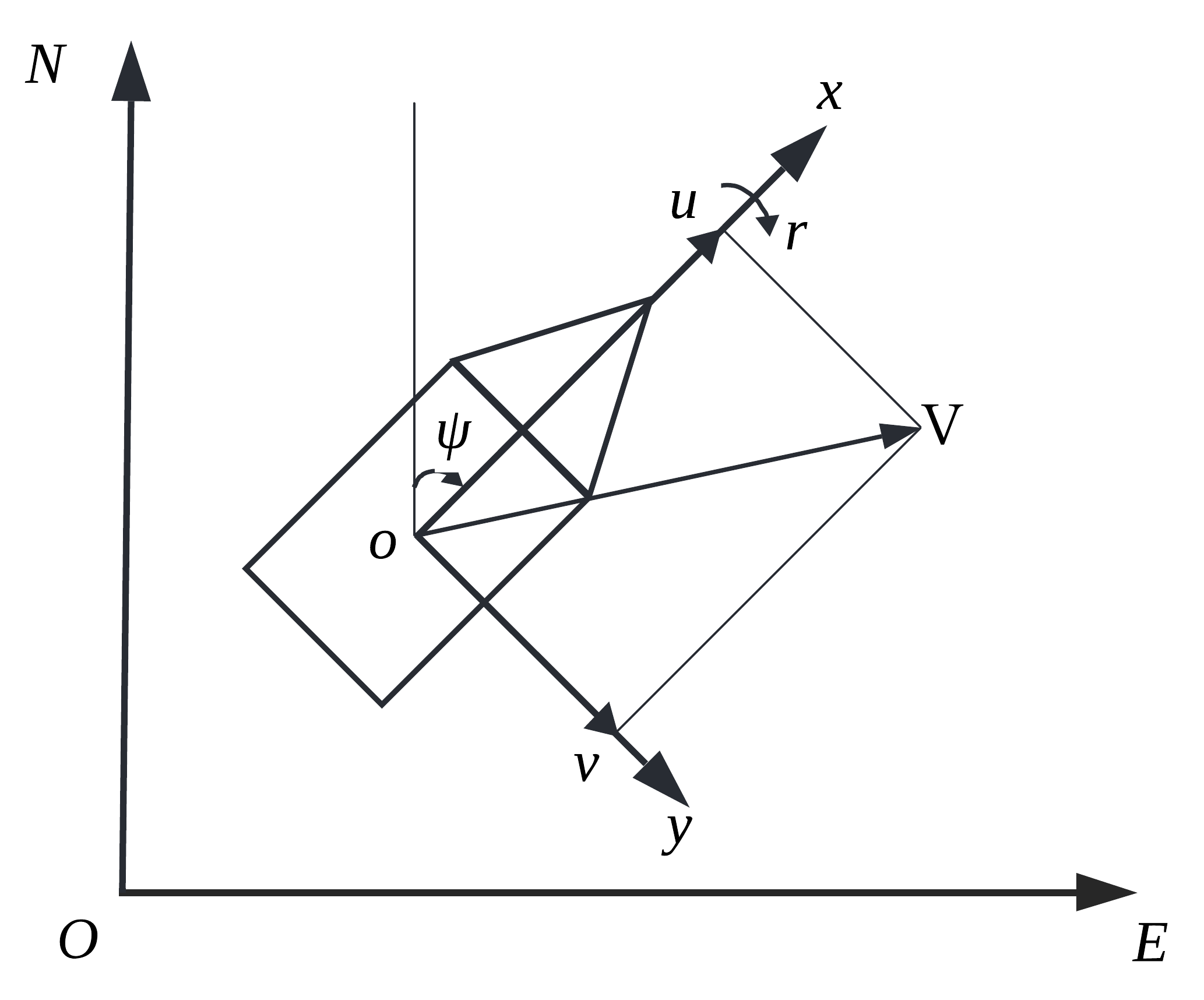}
    \caption{Planar manoeuvring diagram of a \acs{USV} where $NOE$ denotes the geodetic \acs{NED} coordinate system with $O$ as the origin, and $xoy$ denotes the body-fixed coordinate system with $o$ as the \acs{USV}'s centre of gravity. $V$ is the resulting velocity vector of $u$ and $v$.}
    \label{fig:usv_fbddiagram}
\end{figure}

The dynamics of a ship are typically described in six degrees of freedom (\acs{DOF}). For manoeuvring models, it is commonly assumed that the vessel is laterally and longitudinally stable, with negligible roll, pitch, and heave motion \cite{skjetne2004_shipmodeling}. Under these conditions, the model reduces to a 3-\ac{DOF} system described by the position and orientation vector \( \vec{\upeta} = [x, y, \psi]^\intercal \) and the body-fixed velocity vector \( \vec{\upnu} = [u, v, r]^\intercal \), where \((x, y)\) represents the Cartesian position, \(\psi\) is the yaw angle, \((u, v)\) are the surge and sway velocities, and \(r\) is the yaw rate. This manoeuvring model is visualised in \autoref{fig:usv_fbddiagram}.

Accurate modelling of vessel dynamics requires identifying numerous hydrodynamic parameters, where small discrepancies in sensitive values can significantly degrade model predictions and control performance \cite{wang2014_parametersensitivity}. This process is further complicated by environmental disturbances such as wind, waves, and currents. Accurate real-time estimation of such effects typically requires dedicated sensors for measuring free-stream air and water velocities or radar-based techniques for wave spectrum estimation \cite{habashneh2014_spectrumestimationradar}. As a result, optimal control methods have seen limited adoption in \ac{USV} control, and approaches that are more robust to uncertainties are often preferred \cite{haiton2023_reviewpathfollowingcontrol}.






\section{\acl{ADRC}}
This section first introduces the primary design principle behind \ac{ADRC}, followed by an overview of a typical \ac{ADRC} scheme and its essential components.

\subsection{\ac{ADRC} Design Principle}
\label{subsec:adrc_principle}
As outlined in \autoref{sec:preliminaries}, accurately modelling the system or the disturbances acting on it is a significant challenge. This limitation restricts the applicability of model-based control techniques in seagoing \ac{USV}s, which are subject to considerable and often unpredictable disturbances. In contrast, \ac{ADRC} reframes the problem by eliminating the need for precise knowledge of the system dynamics or disturbances. Instead, the unknown dynamics and external disturbances are seen as something to overcome by the control signal \cite{han2009_frompidtoadrc}.

Consider a second-order system represented by the following equations:

\begin{equation}
    \begin{cases}
        \dot{x_1} = x_2 \\
        \dot{x_2} = F(t) + bu \\
        y = x_1 
    \end{cases}
    \label{eq:adrc_basesystem}
\end{equation}

where $y$ is the system output, $u$ represents the input and $F(t) = f(x_1, x_2, w(t), t)$ describes both the system states $x_1$ and $x_2$, and disturbances $w(t)$ as a function of time. Although $F(t)$ may be unknown, the goal is to use the control effort $u$ to compensate for it. In \ac{ADRC}, $F(t)$ is treated as an additional state variable, termed the \enquote{total disturbance} and denoted as $x_3$. This reformulates the original system as:

\begin{equation}
    \begin{cases}
        \dot{x_1} = x_2 \\
        \dot{x_2} = x_3 + bu \\
        \dot{x_3} = G(t) \\
        y = x_1 
    \end{cases}
    \label{eq:adrc_systemformulation}
\end{equation}

where $G(t) = \dot{F}(t)$. 





\subsection{Basic \ac{ADRC} Scheme}
\begin{figure}
    \centering
    \medskip
    \includegraphics[width=1\linewidth]{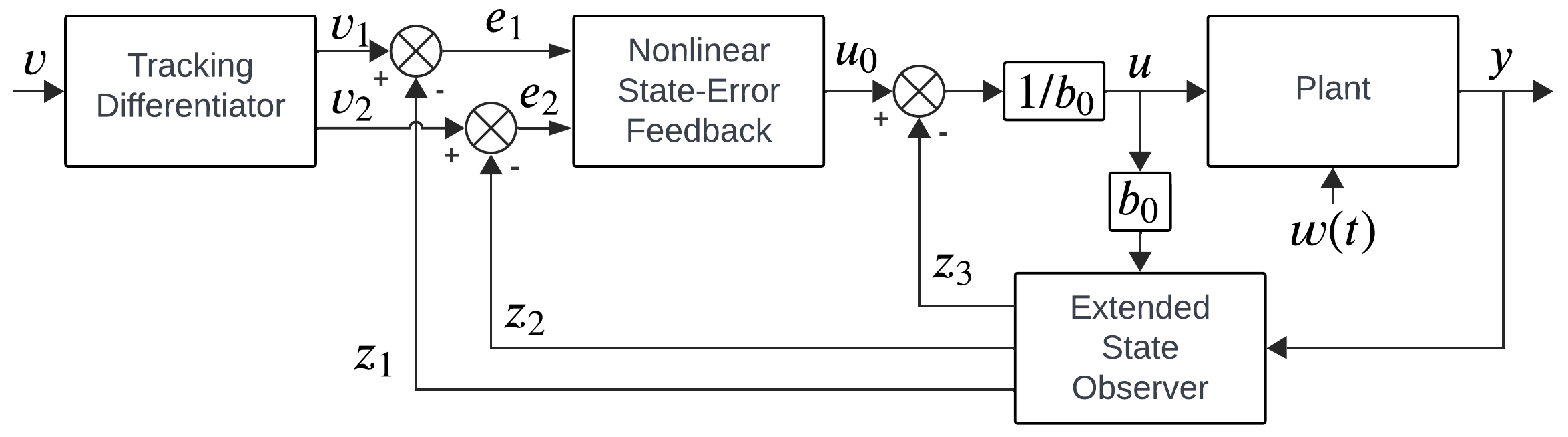}
    \caption{Standard ADRC topology.}
    \label{fig:adrc_topology}
\end{figure}

The standard \ac{ADRC} scheme, as proposed by \cite{han2009_frompidtoadrc}, comprises three primary components. These components are illustrated in \autoref{fig:adrc_topology} and are described in detail in the following subsection.

\subsubsection{Extended State Observer}
A state observer is constructed to estimate the total disturbance $x_3$. This observer, referred to as the \ac{ESO}, is expressed as:

\begin{equation}
    \begin{cases}
        e = z_1 - y \\
        \dot{z_1} = z_2 - \beta_{01} e \\
        \dot{z_2} = z_3 + b_0u -\beta_{02} \operatorname{fal}(e, \alpha_1, \delta) \\
        \dot{z_3} = -\beta_{03} \operatorname{fal}(e, \alpha_2, \delta) \\
    \end{cases}
\end{equation}

where \(\vec{z} =  [z_1, z_2, z_3]^\intercal\) is the state observer estimate of \(\vec{x} =  [x_1, x_2, x_3]^\intercal\), $\beta_{01}\), $\beta_{02}$, and $\beta_{03}$ are the observer gains, and $\operatorname{fal}(e, \alpha, \delta)$ is a nonlinear function that replaces the proportional error $e$ for $z_2$ and $z_3$ \cite{han2009_frompidtoadrc}. The function $\operatorname{fal}(e, \alpha, \delta)$ is defined as:

\begin{equation}
    \operatorname{fal}(e, \alpha, \delta) =
    \begin{cases}
        \frac{e}{\delta^{\alpha-1}}, & \quad |e| \leq \delta \\
        |e|^{\alpha} \operatorname{sign}(e), & \quad |e| > \delta
    \end{cases}
    \label{eq:fal}
\end{equation}

Here, $\delta$ primarily determines the size of the linear region near the origin, while $\alpha$ controls the linear region's and nonlinear region's slope for $|e| > \delta$.  This nonlinear function is designed to improve convergence to the system state while minimising peaking, where the state estimation error temporarily exhibits large transients due to high observer gains in response to sudden changes or disturbances \cite{zhao2016_adrcbook}. The values of $\alpha$ are typically set as $\alpha_1 = 0.5$ and $\alpha_2 = 0.25$ \cite{han2009_frompidtoadrc, Hu2021_ADRCcoursecontrol, chen2015_controlandsimofADRCofUSV}, with $\delta$ remaining a user-defined parameter. 




\subsubsection{Tracking Differentiator}
The \ac{TD} generates a transient profile that the system can reasonably follow to avoid sudden setpoint jumps. The ADRC scheme assumes an underlying second-order system, so a double integral plant can construct this profile. \cite{han2009_frompidtoadrc} proposes a discrete-time solution to such a double integral plant as:

\begin{equation}
    \begin{cases}
        v_1 = v_1 + h v_2 \\
        v_2 = v_2 + h u, & |u| \leq r \\
        u = \operatorname{fhan}(v_1 - v, v_2, r_0, h_0) 
    \end{cases}
\end{equation}

Where $v_1$ and $v_2$ are the transient state and state derivative, respectively, $v$ is the controller setpoint, $r$ is a parameter that can speed up or slow down the transient profile, and $r_0$ and $h_0$ are controller parameters. The discrete time-optimal solution $\operatorname{fhan}(v_1, v_2, r_0, h_0)$ can be written as:

\begin{equation}
    \begin{cases}
        d = r_0 h_0 \\
        d_0 = h_0 d \\
        y = v_1 + h_0 v_2 \\
        a_0 = \sqrt{d^2 + 8 r |y|}\\
        a= 
        \begin{cases}
            v_2 + \frac{a_0 - d}{2} \operatorname{sign}(y), & |y| > d_0 \\
            v_2 + \frac{y}{h}, & |y| \leq d_0 
        \end{cases} \\
        \operatorname{fhan} = -
        \begin{cases}
            r \operatorname{sign}(a), & |a| > d \\
            r \frac{a}{d}, & |a| \leq d
        \end{cases}
    \end{cases}
    \label{eq:fhan_description}
\end{equation}

Per \cite{han2009_frompidtoadrc}, this solution guarantees optimal convergence from $v_1$ to $v$ without overshoot when $r_0 = r$ and $h_0 = h$. However, these parameters can be individually adjusted to change the tracking speed and smoothness of the transient profile, respectively.

\subsubsection{Nonlinear State Error Feedback}
Similar to the \ac{TD}, the basic \ac{ADRC} scheme proposes the use of the optimal solution to the double-integral plant, $\operatorname{fhan}(e_1, ce_2, r_1, h_1)$, as a control law \cite{han2009_frompidtoadrc}:

\begin{equation}
    \begin{cases}
        e_1 = v_1 - z_1 \\
        e_2 = v_2 - z_2 \\
        u_0 = \operatorname{fhan}(e_1, ce_2, r_1, h_1) \\
        u = \frac{u_0 - z_3}{b_0}
    \end{cases}
    \label{eq:adrc_controllaw}
\end{equation}

where $c$ denotes the damping factor, an additional user parameter, and $b_0$ represents the control coefficient, which scales the magnitude of the control signal. This control law assumes that the underlying plant is a second-order system, as presented in \autoref{eq:adrc_systemformulation}, with the total disturbance $z_3$ compensated for via a feed-forward term.


\section{Controller Design}
\begin{figure} 
    \centering 
    \medskip
    \includegraphics[width=\linewidth]{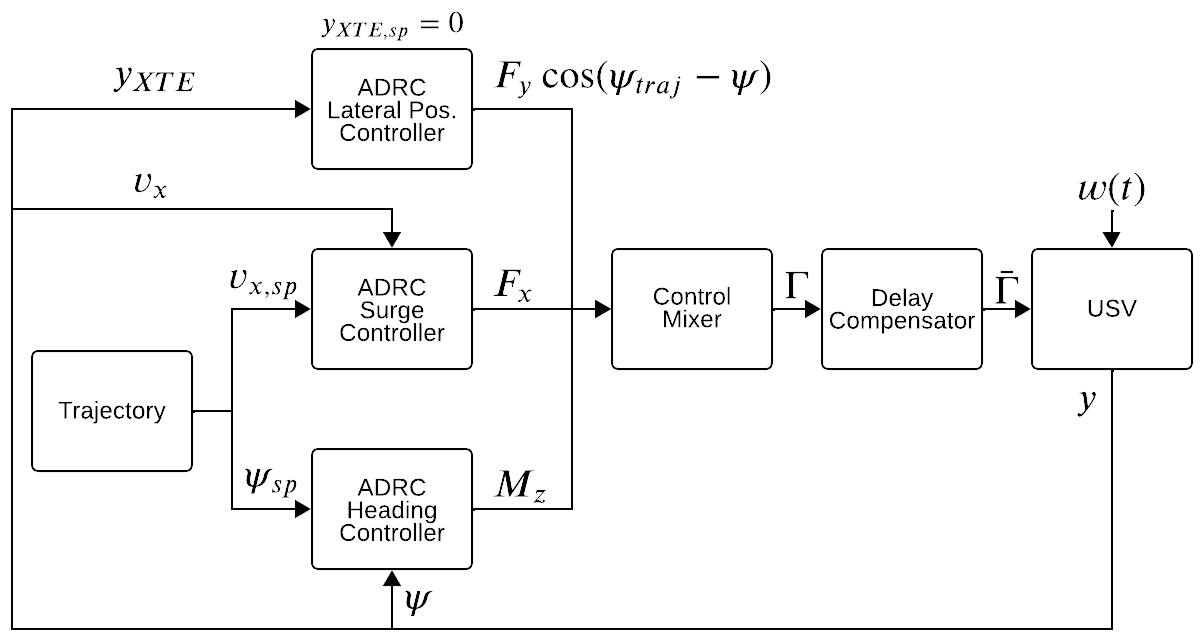} 
    \caption{Design of the trajectory-tracking controller using ADRC topology.} 
    \label{fig:controller_design} 
\end{figure}

The trajectory controller consists of three ADRC controllers operating in parallel, as shown in \autoref{fig:controller_design}, one for each degree of freedom. Each ADRC controller has a topology identical to that shown in \autoref{fig:adrc_topology}. Although the dynamics of each degree of freedom are coupled, the effective decoupling performance of ADRC allows for individual control of each degree of freedom \cite{li2022_dpADRC, zheng2009_decouplingcontrol}.

The heading setpoint is derived from the trajectory using an L1 guidance law, which steers the USV towards the reference trajectory. The surge velocity is also derived from the trajectory and is set to the mission speed of $1.4 \text{m/s}$, automatically reducing during cornering to improve manoeuvrability. The setpoint of the lateral offset is fixed at zero, ensuring that the vessel follows the trajectory line. The output $F_y$ of the ADRC lateral position controller is projected using the vessel's heading $\psi$ and the trajectory line heading $\psi_{traj}$, ensuring $F_y$, and by extension $F_B$, primarily act when the USV is aligned with the reference trajectory.

\subsection{Control Mixer}
The control mixer translates the surge force $F_x$, sway force $F_y$, and yaw moment $M_z$ from the controllers into individual thruster commands. As the system is underactuated, multiple solutions exist; however, under the following assumptions, a unique solution can be derived:

\begin{itemize}
    \item The bow thruster is used exclusively to control the vessel's lateral position, with torque being controlled using the stern thrusters.
    \item The stern thrusters, mounted at an inward angle of $\alpha$, have a negligible lateral force contribution ($\sin{\alpha} \approx 0$).
\end{itemize}

Then, the control mixer is defined as:

\begin{equation}
    \begin{cases}
        F_{SL} = \frac{1}{2\cos{\alpha}} \left( F_x + \frac{M_z + F_y x_{BT}}{y_{ST}} \right) \\
        F_{SR} = \frac{1}{2\cos{\alpha}} \left( F_x - \frac{M_z + F_y x_{BT}}{y_{ST}} \right) \\
        F_{B} = F_y
    \end{cases}
\label{eq:controlmixer}
\end{equation}

where $F_{SL}$, $F_{SR}$, and $F_{B}$ represent the thrusts of the left stern thruster, right stern thruster, and bow thruster, respectively. The parameters $x_{BT}$ and $y_{ST}$ are geometric dimensions of the USV, denoting the longitudinal distance from the centre of mass to the bow thruster and the lateral distance to the stern thrusters, respectively.

\subsection{Delay Compensator}
Thruster speed changes are subject to spin-up and spin-down delays due to rotational inertia and propeller drag. A first-order low-pass filter can approximate this delay \cite{zou2018adrc_quadrotor}:

\begin{equation}
    \Omega(s) = \frac{\Omega_0(s)}{1 + s \tau_d}
\end{equation}

where \( \Omega(s) \) denotes the actual thruster speed, \( \Omega_0(s) \) is the setpoint, and \( \tau_d \) represents the time constant, approximated as \( \tau_d = T_s / 4 \) \cite{nise2020control}.

For the \acs{DUS} V2500, settling times were identified as approximately \( T_s \approx 2 \) seconds for the stern thrusters and \( T_s \approx 1 \) second for the bow thruster. Although not exact, this linear approximation suffices for delay compensation in \ac{ADRC} \cite{han2009_frompidtoadrc, zou2018adrc_quadrotor}.

The delay compensator leverages the first-order approximation of the motor delay. For the total control signal $\Vec{\Gamma} = [F_{SL}, F_{SR}, F_B]^\intercal$, the delay is compensated using the following first-order approximation:

\begin{equation}
    \bar{\Vec{\Gamma}} = \Vec{\Gamma} + \Vec{\uptau_d} \dot{\Vec{\Gamma}}
\end{equation}

where $\bar{\Vec{\Gamma}}$ represents the compensated control signal, $\dot{\Vec{\Gamma}}$ is the derivative of the control signal, and $\Vec{\uptau_d} = [0.5, 0.5, 0.25]^\intercal$ contains the time constants for the stern and bow thrusters.

The derivative $\dot{\Vec{\Gamma}}$ is calculated using a separate \acf{TD} for each control signal. By setting a large value for $r_0$, the transient profile of the \ac{TD} instantly adapts to changes in the input signal, effectively acting as a differentiation filter. This approach offers improved noise tolerance compared to numerical differentiation methods \cite{wang2020_comparisonTD}.


\section{Simulation Design}
\begin{figure}[b]
    \centering
    \includegraphics[width=\linewidth]{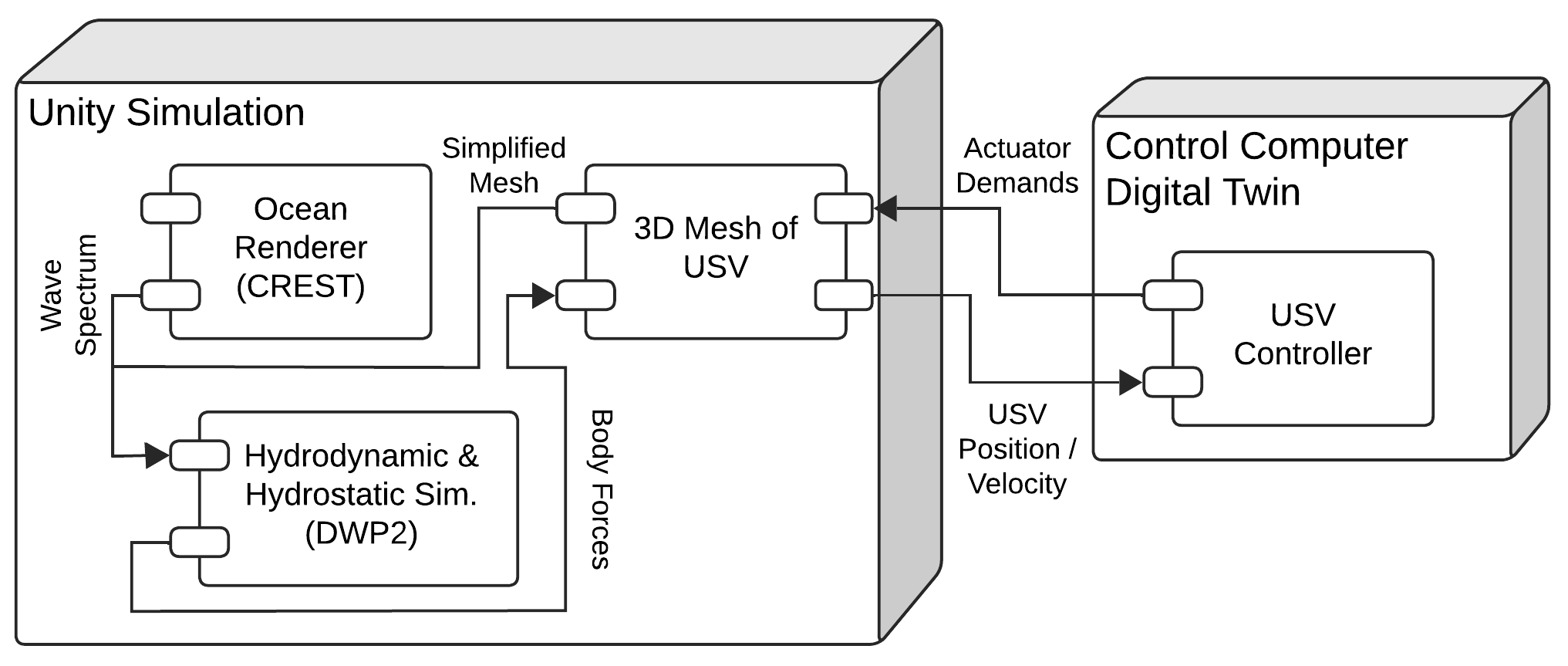}
    \caption{Architecture of Unity Simulation.}
    \label{fig:unitysim_uml}
\end{figure}

A \ac{SITL} simulation is used to validate the controller, comprising a digital twin of the onboard control computer integrated with a Unity-based simulation of vessel dynamics and sensor inputs. This simulation architecture, shown in \autoref{fig:unitysim_uml}, provides an environment for validating and evaluating controller performance under realistic operating conditions. An example of a USV in this simulation environment can be seen in \autoref{fig:DUS_V2500}.

\begin{figure}
    \centering
    \medskip
    \includegraphics[width=\linewidth]{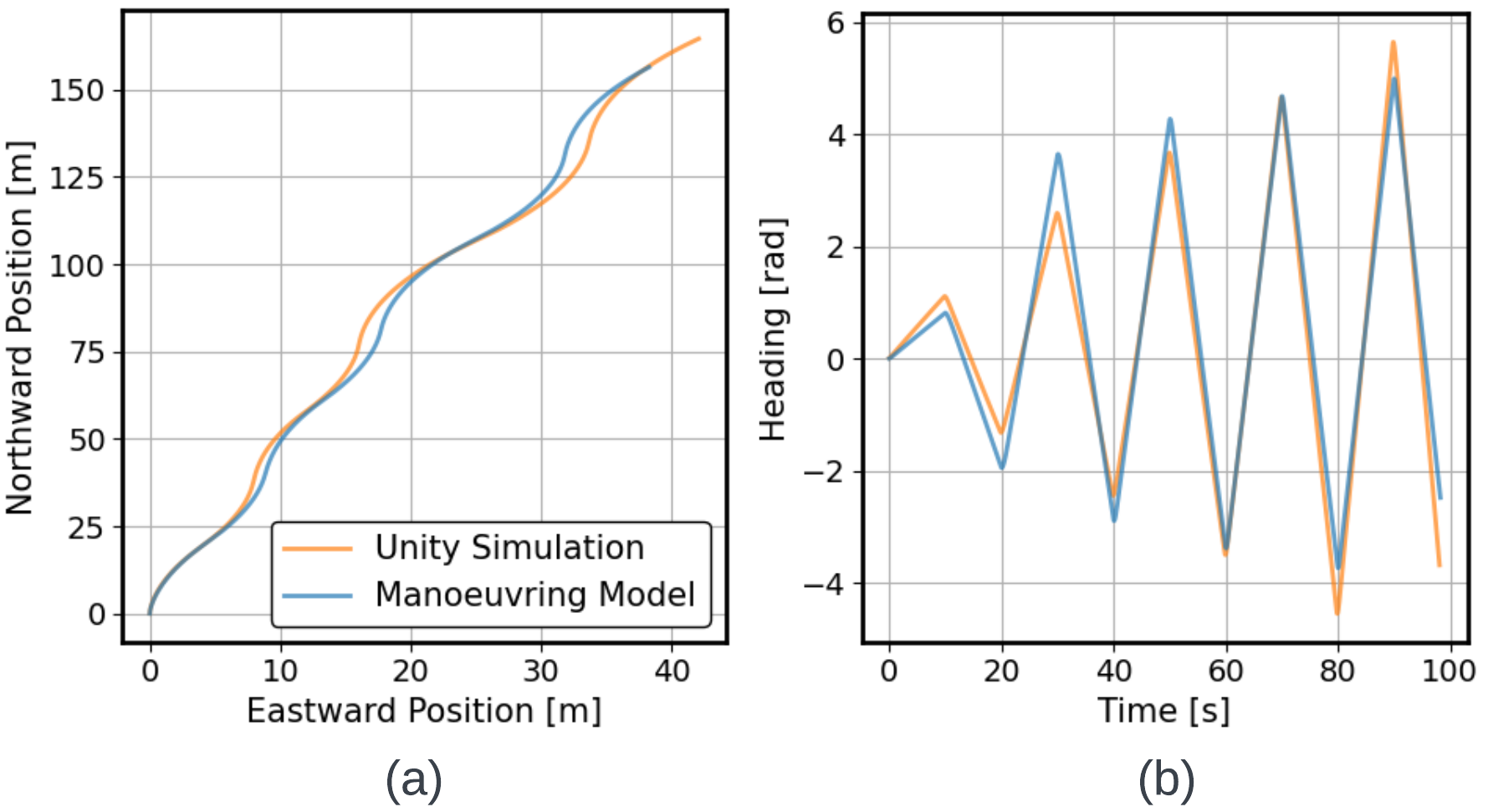}
    \caption{Verification of the Unity simulation compared to a known manoeuvring model: (a) Zig-zag manoeuvre and (b) In-place spin manoeuvre. The same thrust command sequences are fed to the vessel in our Unity simulator and an analytic manoeuvring model with parameters identified from real-world data.}
    \label{fig:sim_verification}
\end{figure}

This Unity simulation uses the \ac{DWP2} package to model vessel dynamics in six \ac{DOF} \cite{DWP2}. Although a $3 \text{DOF}$ manoeuvring model of the V2500 is available, based on system identification data, realistic simulation of environmental disturbances requires including roll, pitch, and heave dynamics. The \ac{DWP2} physics simulation is verified against an analytic manoeuvring model of the \ac{DUS} V2500 with parameters identified from real-world data, as shown in \autoref{fig:sim_verification}. However, verification of the additional degrees of freedom (roll, pitch, and heave), which are primarily influenced by wave disturbances, is limited due to the absence of relevant system identification data or models. Collecting such data is costly, requiring wave tank experiments to establish known and precise wave conditions. 

To simulate a realistic wave environment, \ac{DWP2} is integrated with the open-source CREST 4 ocean renderer, which generates a Pierson-Moskowitz wave spectrum \cite{crest}. The desired intensity of the wave spectrum can be easily specified to determine the total wave height, with sea states corresponding to Douglas Sea States 1-4 implemented within the simulation. This corresponds to a maximum wave height of $2.5 \text{m}$ at sea state 4. Furthermore, CREST 4 supports fluid flow modelling, allowing the inclusion of ocean currents in the simulation. 


\section{Simulated Experiment}
\label{sec:simexperiment_design}
Simulated experiments are conducted to evaluate the performance of the \ac{ADRC} controller against a baseline PID controller in a controlled environment. The experiment uses a predefined trajectory within the Derde Haven of Scheveningen, created with the navigation software stack developed by \ac{DUS}. This software enables waypoint-defined trajectories that are compatible with both simulations and real-world testing.

\begin{figure}
    \centering
    \medskip
    \includegraphics[width=0.55\linewidth]{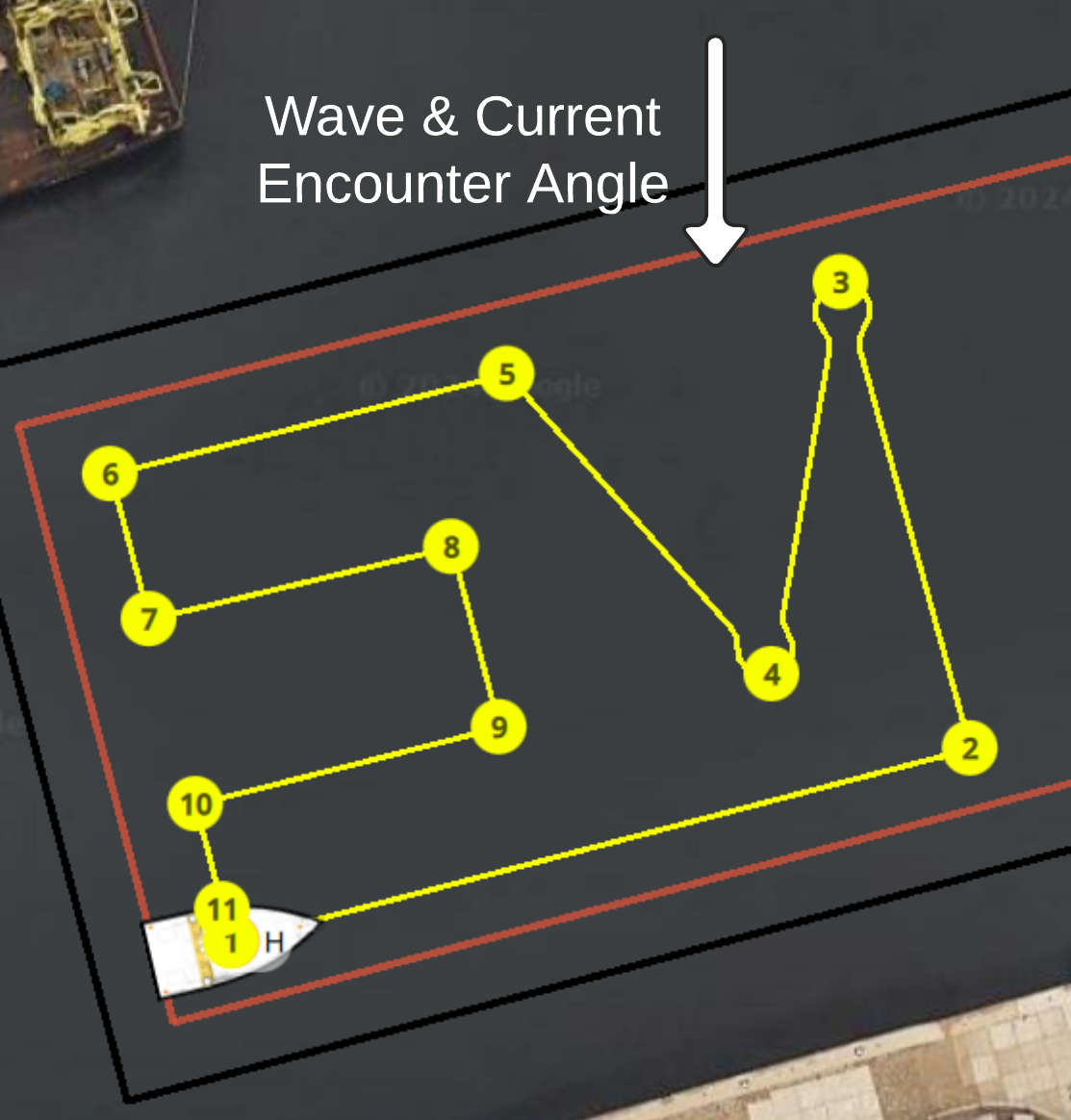}
    \caption{In-harbour trajectory used for both simulated and real-world tests. The encounter angle for waves and currents applies only to the simulated experiment.}
    \label{fig:trajectory_map}
\end{figure}

The trajectory of the simulated experiment, shown in \autoref{fig:trajectory_map}, consists of a combination of straight paths and corners, including acute turns and 90 degrees. Simulated disturbances, including waves and currents, originate from the north, as indicated by the white arrow in the figure. The trajectory is defined with a mission speed of $1.4 \, \text{m/s}$. Dubins curves are used with a turning radius derived from the mission speed to ensure smooth transitions between segments. In simulation, the turning radius is set smaller than in real-world operations, as the physical limitations of the vessel are less restrictive. This adjustment enables the evaluation of more challenging trajectories, highlighting differences in controller performance more effectively.

Performance of both the \ac{ADRC} controller and the baseline PID controller -- currently deployed on the \ac{DUS} V2500 -- are evaluated using the same trajectory. The PID controller uses the gain values implemented in the production vessel, while the gains of the \ac{ADRC} controller were manually tuned for the experiment. Both controllers follow the control architecture depicted in \autoref{fig:controller_design}, with the baseline replacing the \ac{ADRC} controller with a standard PID controller.

Each controller is assessed in trajectory tracking under four distinct operating conditions: 

\begin{enumerate}
    \item No disturbances.
    \item A current of $0.5 \text{m/s}$ with no waves.
    \item No current with sea state 4 waves ($2.5 \text{m}$ wave height).
    \item Both a $0.5 \text{m/s}$ current and sea state 4 waves.
\end{enumerate}

The performance of the controllers is evaluated based on the following metrics:

\begin{itemize}
    \item \textbf{\ac{XTE}:} The deviation from the desired trajectory.
    \item \textbf{Total Battery Usage:} The total battery capacity consumed (in Ampere-hours) by the USV, estimated by integrating the current delivered to each motor over the entire trajectory.
\end{itemize}

\subsection{Results}
\begin{figure}[b]
    \centering
    \includegraphics[width=0.9\linewidth]{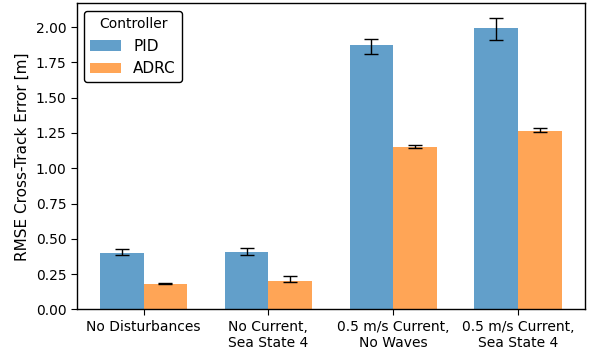}
    \caption{\ac{RMS} of the cross-track error (\ac{XTE}) for each controller, averaged over 5 simulated trajectories. Error bars indicate min- and maximum values.}
    \label{fig:XTE_simulation}
\end{figure}

\begin{figure}
\medskip
    \centering
    \includegraphics[width=0.9\linewidth]{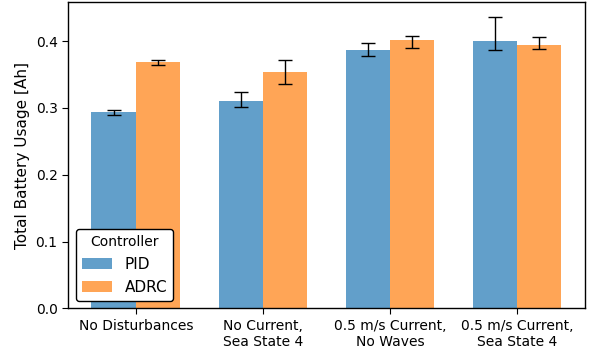}
    \caption{Total battery consumption of each controller, averaged over 5 simulated trajectories. Error bars indicate min- and maximum measured values.}
    \label{fig:batteryusage_simulation}
\end{figure}

To evaluate the performance of each controller under varying conditions, five repetitions of the trajectory are conducted for each case to account for error margins. The \ac{RMS} of the \ac{XTE} is computed over the length of each trajectory. This metric for each controller and condition is summarised in \autoref{fig:XTE_simulation}. Additionally, the total energy usage for each controller, averaged over 5 trajectories for each testing condition, is presented in \autoref{fig:batteryusage_simulation}. 

\begin{figure*}
    \centering
    \medskip
    \includegraphics[width=\linewidth]{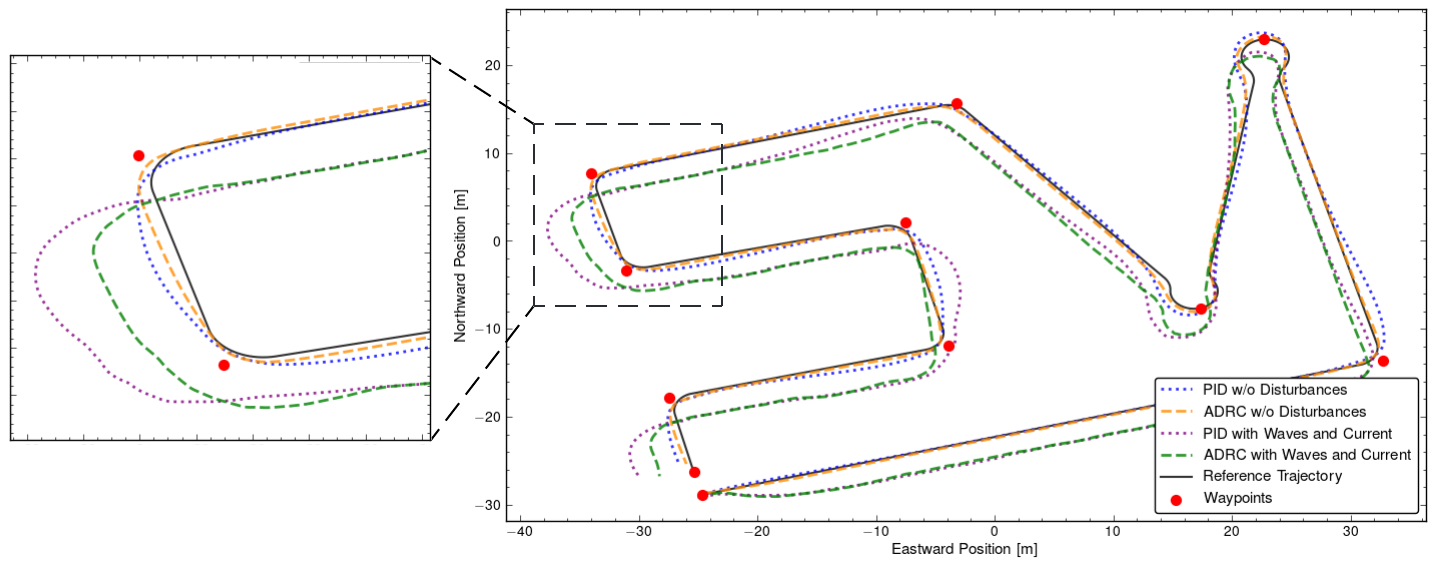}
    \caption{Comparison of \ac{ADRC} and \ac{PID} controllers in simulation under different conditions. The steady-state error for both \ac{ADRC} and \ac{PID} in the presence of currents is due to input saturation on the bow thruster (see \autoref{eq:controlmixer} and \autoref{sec:sim_discussion}).}
    \label{fig:simtrajectory}
\end{figure*}

A sample of the trajectories for both the \ac{ADRC} and \ac{PID} controllers, under conditions without disturbances and with all disturbances active ($0.5 \text{m/s}$ current and sea state 4), is shown in \autoref{fig:simtrajectory}. 

\subsection{Discussion}
\label{sec:sim_discussion}
The results demonstrate a clear performance improvement of the \ac{ADRC} controller over the \ac{PID} controller under all conditions. The ADRC controller achieves significant reductions in \ac{XTE}, even in the absence of disturbances, indicating that the improved trajectory tracking is primarily due to the \ac{ADRC} control law (\autoref{eq:adrc_controllaw}) rather than its disturbance rejection capabilities.

As shown in \autoref{fig:simtrajectory}, both controllers exhibit a constant lateral offset during straight-line tracking with currents. This offset results from the guidance law striving to align the vessel with the path. Only the bow thruster can compensate for the lateral offset in this configuration, but as it is not strong enough, its input saturates. Alternative guidance laws could also encourage using stern thrusters to correct the lateral offset, mitigating this effect.

Wave-induced disturbances have minimal impact on lateral error, with \ac{XTE} remaining similar to no-wave conditions. Although slight trajectory variability is observed, it did not significantly affect overall tracking performance. This is likely due to limitations in the wave model, which may not fully capture wave-to-vessel momentum transfer. Despite increased pitch, roll, and heave motions, vessel manoeuvrability remained largely unaffected.

Battery consumption, illustrated in \autoref{fig:batteryusage_simulation}, is higher for the \ac{ADRC} controller in calm conditions but comparable to the \ac{PID} controller when exposed to currents. In the presence of current, \ac{ADRC} demonstrates lower consumption by completing the trajectory quicker, as shown in \autoref{fig:simtrajectory}.

The increased control effort of \ac{ADRC} arises from the fundamental principle of counteracting \(F(t) = f(x_1, x_2, w(t), t)\) entirely through the control signal \(u\), as defined in \autoref{eq:adrc_systemformulation}. This forces the system to behave as a second-order system, actively rejecting external disturbances and internal higher-order effects. However, since \ac{ADRC} cannot distinguish between internal dynamics and external disturbances, its energy efficiency is inherently reduced.



\section{Real-World Experiment}

\begin{figure}
    \centering
    \includegraphics[width=\linewidth]{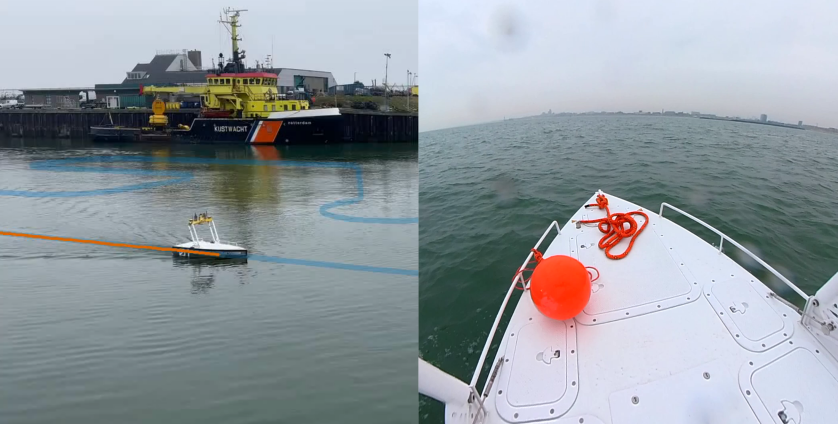}
    \caption{The DUS V2500 during in-harbour trials (left) and during near-shore trials (right) in Scheveningen, the Netherlands.}
    \label{fig:dusv2500_experiments}
\end{figure}

Field trials are conducted in two scenarios to evaluate the \ac{ADRC} controller under varying environmental conditions. The first scenario involves replicating the trajectory, defined by the same waypoints as in \autoref{fig:simtrajectory}, within the controlled environment of Scheveningen Harbour, with minimal disturbances, as shown in \autoref{fig:dusv2500_experiments}.

The second scenario tests an identical trajectory in a near-shore environment outside the harbour to assess performance under increased wave and current disturbances, approximately 600 metres from the nearest groyne and 1 kilometre from the shore. An on-board view is shown in \autoref{fig:dusv2500_experiments}.

The gains and parameters for both controllers are kept identical to those used in the simulation, as they demonstrated satisfactory performance during field tests. This approach ensures a fair comparison between the controllers and highlights their ability to transfer effectively from simulation to real-world scenarios.

\subsection{Results}
The weather conditions during the experiments are summarised in \autoref{tab:experiment_conditions} and are based on meteorological data. Local conditions may vary slightly, but could not be recorded due to the absence of required sensors. In-harbour tests are conducted under minimal disturbance, whereas sea trials are conducted in relatively calm weather. More challenging conditions were not available during our field experiments.

\definecolor{Silver}{rgb}{0.752,0.752,0.752}
\definecolor{GreenSpring}{rgb}{0.729,0.741,0.713}
\begin{table}[b]
\caption{Experimental conditions of field trials.}
\label{tab:experiment_conditions}
\centering
\begin{tblr}{
  row{even} = {c},
  row{1} = {Silver},
  row{3} = {c},
  row{5} = {c},
  row{7} = {c},
  cell{1}{1} = {c},
  cell{1}{2} = {c},
  cell{1}{3} = {c},
  cell{1}{4} = {GreenSpring},
  cell{2}{1} = {r=3}{},
  cell{5}{1} = {r=3}{},
  vlines,
  hline{1-2,5,8} = {-}{},
  hline{3-4,6-7} = {2-4}{dashed},
}
\textbf{Experiment} & \textbf{Condition} & \textbf{Magnitude} & \textbf{Direction} \\
\textit{In Harbour} & Sea State          & 0-1                & N/A                \\
                    & Wind Speed         & 6 kts              & SSW                \\
                    & Current Speed      & N/A                & N/A                \\
\textit{At Sea}     & Sea State          & 1-2                  & N/A                \\
                    & Wind Speed         & 8 kts              & SSE                \\
                    & Current Speed      & 0.7 kts            & SSW                
\end{tblr}
\end{table}

\begin{figure}
    \centering
    \includegraphics[width=\linewidth]{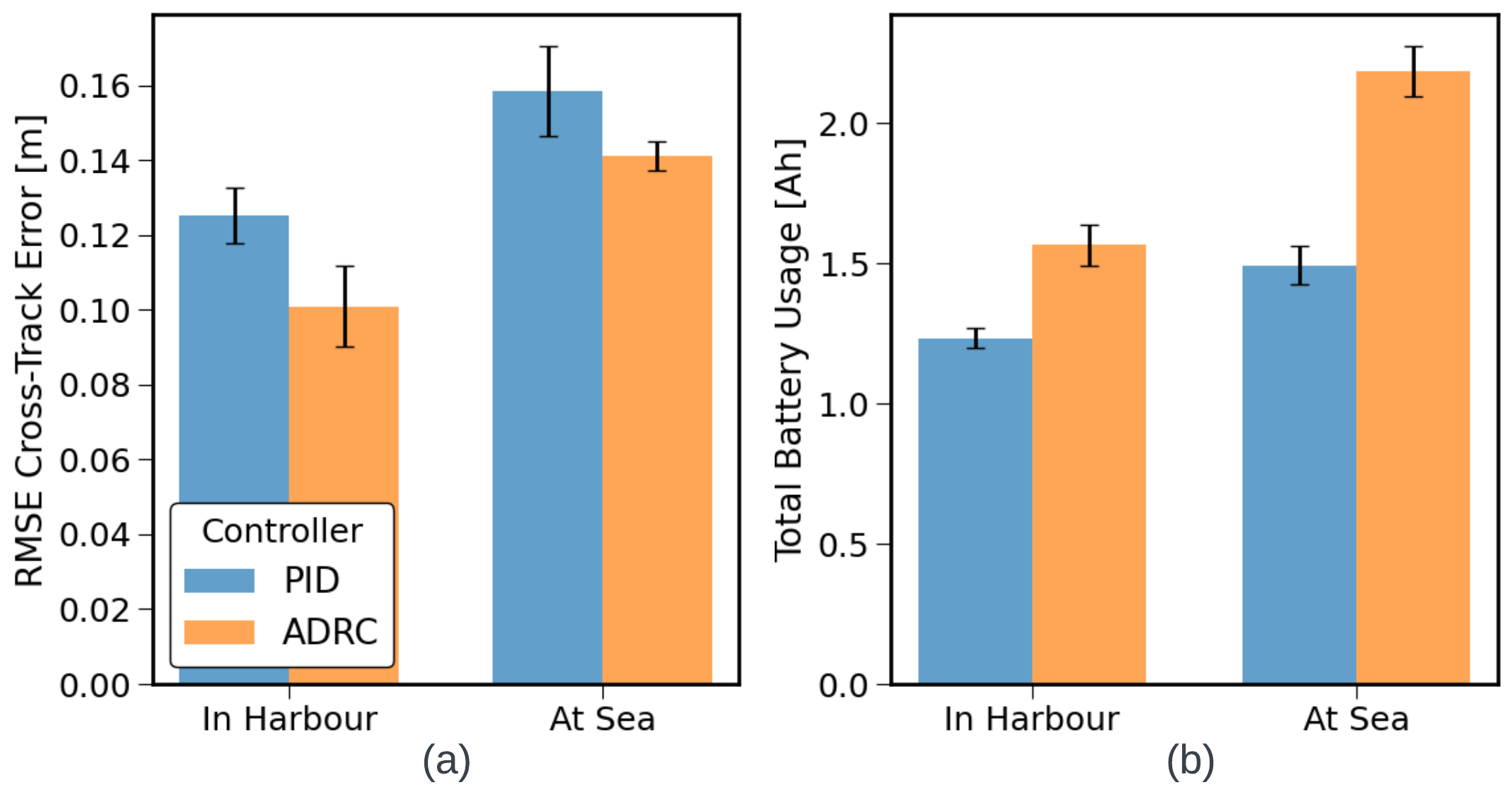}
    \caption{(a) \ac{RMS} of the cross-track error for each controller. (b) Total battery consumption of each controller. Both metrics are averaged over two measured trajectories. Error bars indicate the minimum and maximum measured values.}
    \label{fig:real_results}
\end{figure}

Identical metrics used in the simulation trials are applied to the field trials. Both the \ac{PID} and \ac{ADRC} controllers are evaluated on trajectories performed in the harbour and at sea. For each controller and location, two trajectories were recorded, resulting in a total of eight trajectories. The results, summarised in \autoref{fig:real_results}, present the \ac{XTE} and the total battery usage. Both metrics were averaged over the two recorded trajectories.


\subsection{Discussion}
Real-world tests show similar trends to simulation: as seen in \autoref{fig:real_results}, \ac{ADRC} reduces \ac{XTE} under all conditions, albeit with higher energy use. However, the XTE improvement is less pronounced than in simulation. As noted in \autoref{sec:simexperiment_design}, safety constraints required wider turns, likely reducing ADRC’s relative advantage.

A comparison of key outcomes is as follows:
\vspace{0.05cm}
\subsubsection{Similarities}
\begin{itemize}
    \item \ac{ADRC} consistently outperforms \ac{PID} in reducing \ac{XTE}.
    \item Energy use is higher for \ac{ADRC} in mild conditions but comparable under stronger currents.
    \item Bow thruster saturation limits lateral correction in both domains.
\end{itemize}
\subsubsection{Discrepancies}
\begin{figure}
    \centering
    \includegraphics[width=\linewidth]{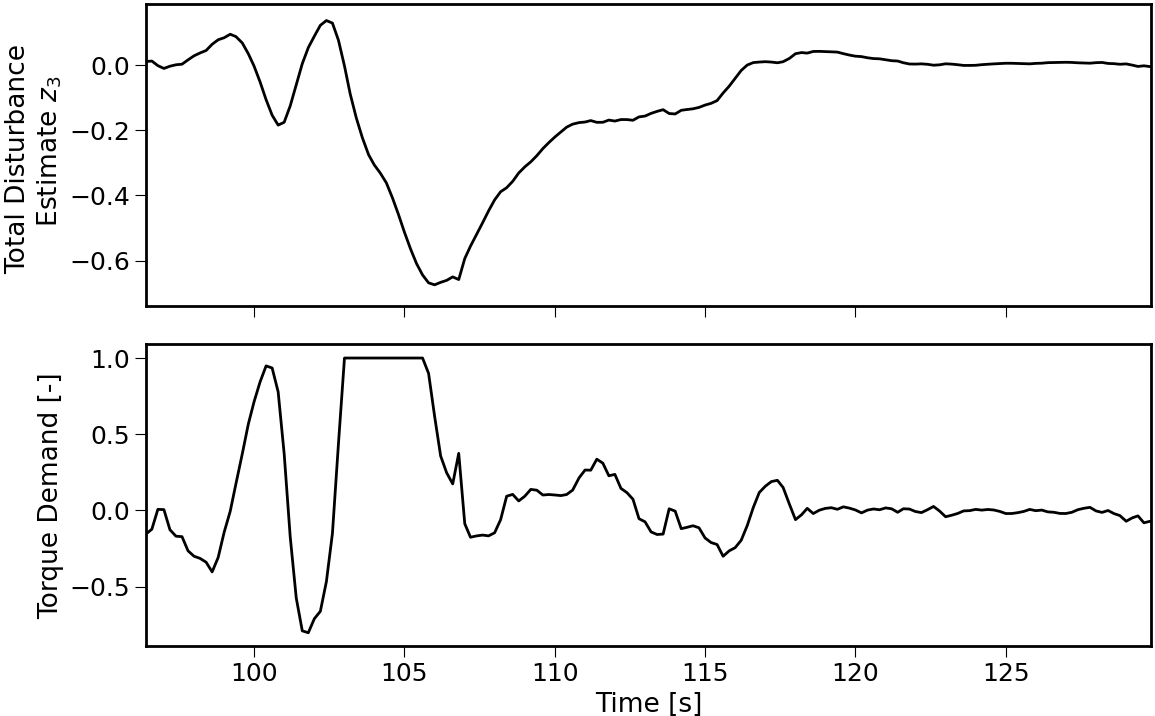}
    \caption{Spike in the total disturbance estimate $z_3$ during near-shore trials, resulting in torque saturation and increased power consumption.}
    \label{fig:disturb_est_spike}
\end{figure}

\begin{itemize}
    \item XTE reduction is smaller in the field ($10$–$20\%$) than in simulation ($30$–$40\%$).
    \item Disturbance estimates $z_3$ spike at sea (\autoref{fig:disturb_est_spike}), unlike in simulation.
    \item Roll/pitch/heave impacts are unclear in field tests due to limited sensor data.
\end{itemize}
\vspace{0.05cm}

These discrepancies suggest that certain dynamics, such as complex wave–vessel interactions, are not fully captured in simulation. The wave model likely underestimates momentum transfer, and the lack of real-time current or wave direction sensing on the DUS V2500 limits real-time disturbance estimation. Although field tests confirm improved path-following performance of \ac{ADRC} over \ac{PID}, they also show that simulation-based evaluation is constrained by these differences. Regardless, field testing results clearly show the trade-off between tracking performance and increased energy consumption of \ac{ADRC}.



\section{Conclusion}
This study evaluates the performance of the \ac{ADRC} framework for \ac{USV} trajectory tracking through simulations and field trials. The results indicate that \ac{ADRC} outperforms \ac{PID} in reducing cross-track error in both simulation (by $30$-$40$\%) and field trials (by $10$-$20$\%).

In addition to the reduction in cross-track error, \ac{ADRC} exhibits increased energy consumption compared to \ac{PID}. In simulation, this increase is observed only in the absence of current. With currents present, \ac{ADRC} completed the trajectory faster, resulting in equal battery usage. In field trials, the transition from harbour to near-shore conditions exacerbates this effect, with \ac{ADRC} consuming approximately $50$\% more energy than \ac{PID}.

In conclusion, while the \ac{ADRC} controller enhances trajectory tracking, its increased energy consumption limits practical use in industrial \ac{USV} applications. Future work should focus on optimising disturbance rejection by scaling the feedforward term and incorporating known system dynamics to better distinguish internal and external effects, thereby improving efficiency.

\appendices

\section*{Acknowledgment}
This study would not have been possible without the generosity of \acl{DUS}, who provided their time and access to existing systems and software and made the V2500 available for testing in Scheveningen. A special thanks to the entire team for their invaluable support in preparing the vessel and assisting during the trials.

\bibliographystyle{IEEEtran} 
\bibliography{bibtex/bib/main}

\end{document}